\documentclass[11pt,letterpaper]{article}
\usepackage{naaclhlt2016}
\usepackage{times}
\usepackage{latexsym}
\usepackage{enumitem}

\usepackage{arydshln}
\naaclfinalcopy 
\def\naaclpaperid{***} 

\title{CLaC at SemEval-2016 Task 11: Exploring linguistic and psycho-linguistic Features for Complex Word Identification}

\author{Elnaz Davoodi \and Leila Kosseim \\
	   Concordia University\\
	   Department of Computer Science and Software Engineering\\
	    Montr{\'e}al, Qu{\'e}bec, Canada H3G 2W1\\
	    {\tt \{{e\_davoo, kosseim}\}@encs.concordia.ca}}

\date{}

\begin{document}
\maketitle

\begin{abstract}
 This paper describes the system deployed by the CLaC-EDLK team to the \textit{SemEval 2016, Complex Word Identification task}. The goal of the task is to identify if a given word in a given context is \textit{simple} or \textit{complex}. Our system relies on linguistic features and cognitive complexity. We used several supervised models, however the Random Forest model outperformed the others. Overall our best configuration achieved a G-score of 68.8\% in the task, ranking our system 21 out of 45.     
\end{abstract}

\section{Introduction}
Text simplification involves reducing the complexity of a text in order to make it more accessible to a larger audience or to a specific audience such as children, second language learners, the elderly, etc. 

Research efforts on text simplification have mostly focused on either lexical \cite{devlin1998,carroll1998,biran2011} or syntactic simplification \cite{chandrasekar1997,siddharthan2006,kauchak2013}. Lexical simplification involves replacing specific words in order to reduce lexical complexity, while still conveying the same information. Lexical simplification is still a challenging task as identifying and simplifying complex words in a given context is not straightforward. Complex word identification is the first step towards lexical simplification. 

For the challenge, we experimented with a number of features that we suspected would be correlated with a word's complexity level and selected five that were the most discriminating on the given training data set. Using a Random Forest classifier, we built two models that we deployed to the Sem-Eval 2016 Complex Word Identification Task. These models achieved better than average ranking compared to the other participating systems. We believe that the approach proposed in this paper can link linguistic features and cognitive complexity together in order to improve lexical simplification.


\section{Complex Word Identification Task Setup}
The SemEval 2016 Complex Word Identification Task required participating systems to automatically classify words as \textit{simple} or \textit{complex} in a given context. For training, participants were given 2247 instances. Each instance consisted of a target word $W$, its offset in the sentence and a class label (0 if $W$ was considered \textit{simple} in the sentence, or 1 if $W$ was \textit{complex} in the sentence). For example, given the training instance 1 below, the target word \textit{happened} is classified as simple; however in training instance 2, the same word is classified as complex.

\textbf{Training instance 1: }There are several stories about Mozart 's final illness and death , and it is not easy to be sure what happened . happened~~21~~0

\textbf{Training instance 2: }Although anoxic events have not happened for millions of years , the geological record shows that they happened many times in the past .  happened~~5~~1

Given the above, the task was to identify if a word is \textit{complex} or \textit{simple} in a test set of 88,221 instances.

\section{Methodology}
We experimented with various supervised models for this binary classification task; however Random Forest outperformed the others. We also experimented with various features and retained the five features described below.

\subsection{Feature Set}
In our final complex word identification system, we used as features: the frequency of the target word in the Web1T Google N-gram Corpus, its Part of Speech (POS) tag, the number of synonyms it has in WordNet, the inverse of its length and a psycholinguistic feature indicating its ``abstract" level. These features were selected based on their discriminating power on the training data. 

\subsubsection{Frequency in the Web1T Google N-gram Corpus}
The Google N-gram corpus \cite{michel2011} is a collection of English one- to five-grams with their frequencies from different sources and from different years. This corpus contains approximately 1 trillion words from the Web. In order to focus on the more recent usage of the words and reduce the size of the corpus, we considered the frequency of the target words in sources which were indexed after year 2000. This way, we reduced the influence of frequent but obsolete words. We used this feature based on the assumption that simpler words are more frequently used.

\subsubsection{Part of Speech Tag}
We suspected that a word's POS tag may influence its complexity level. As a word may be tagged with different parts-of-speech, we suspected that a particular usage may be more complex than another depending on how common that usage is.  For example, the word {\em highlight} may be used as a noun or, less frequently, as a verb.  Hence, {\em highlight} as noun may be more likely to be considered {\em simple} and {\em highlight} as verb may be more likely to be considered {\em complex}.   As a feature, we used the POS tags of target words given in the training and test data sets. The words in each sentence are tagged using Stanford POS tagger \cite{toutanova2003}.

\subsubsection{Number of Synonyms in WordNet}
Another linguistic feature that we suspected may be correlated with the complexity of a word is the number of synonyms it has. Based on our statistical analysis of the training set, complex words have fewer synonyms than simpler words. For instance, the probability that a complex word has less than 4 synonyms is 33.65\% on the training set; while this number is 24.10\% for simple words. This can be explained by the fact that complex words tend to denote specific entities or concepts and therefore tend to have less synonyms. Thus, we considered the number of synonyms of the target word as one of our features. 

\subsubsection{Inverse of Word Length}
Based on the work of traditional text complexity measures such as the Flesch index \cite{kincaid1975}, we took into account the length of a word as a feature to determine its simplicity level. 

\subsubsection{Psycholinguistic Feature}
We suspected that the more abstract a word, the more complex it will be perceived. To investigate the correlation between the degree of abstractness of a word and its complexity, we used the MRC psycholinguistic database \cite{wilson1988}. This electronic resource contains the score of 26 psycholinguistic features for 150,834 words. One such feature indicates the level of abstraction associated with the entity or concept denoted by the word. This concreteness feature is available for 8,228 words and is indicated by an integer value ranging from 100 (very abstract) to 700 (very concrete). For this psycholinguistic feature, if the target word had a concreteness value in the MRC, we used its value as a feature. However, using this feature has a drawback since it does not cover all the words. For now, we deal with this problem by considering the value of out-of-database words as 0.

\begin{table*}[ht]
\small
\centering
    \begin{tabular}{|l|c|} \hline
         \textbf{Feature}& \textbf{ Information Gain} \\ \hline \hline
        
         Number of Synonyms in WordNet &0.048\\
    Psycholinguistic Feature &0.024\\
 Part of Speech Tag &0.015\\
Frequency in Web1T Google N-gram Corpus &0.007\\ 
Inverse of Word Length &0.006\\

   \hline

    \end{tabular}
    \caption{Information gain of each feature.}
    \label{infoGain}
\end{table*}

\begin{table*}[ht]
\centering
    \begin{tabular}{|r|r|r|r|r|} \hline
         \textbf{Learning Model} & \textbf{Precision} & \textbf{Recall} & \textbf{F-Measure}& \textbf{Class} \\ \hline \hline\
         Na{\"i}ve Bayes & 0.680 & 0.978 & 0.803 & 0 (Simple) \\
         Na{\"i}ve Bayes & 0.267 & 0.017 & 0.032 & 1 (Complex)\\
         Na{\"i}ve Bayes & 0.548 & 0.672 & 0.557 & Weighted Average\\ \hdashline
         Neural Network & 0.707 & 0.913 & 0.797 & 0 (Simple)\\
         Neural Network & 0.509 & 0.193 & 0.280 & 1 (Complex) \\ 
         Neural Network & 0.644 & 0.684 & 0.632 & Weighted Average\\ \hdashline
          Decision Tree (J48) & 0.715 & 0.919 & 0.804 & 0 (Simple)\\
         Decision Tree (J48) & 0.551 & 0.212 & 0.306 & 1 (Complex)\\
         Decision Tree (J48) & 0.662 & 0.694 & 0.645 & Weighted Average\\ \hdashline
         Random Forest & 0.738 & 0.815 & 0.775 & 0 (Simple)\\
         Random Forest & 0.491 & 0.383 & 0.430 & 1 (Complex)\\
         Random Forest & 0.660 & 0.677 & 0.667 & Weighted Average \\ \hline

    \end{tabular}
    \caption{Performance of various learning models evaluated using 10-fold cross-validation on the training set.}
    \label{ClassifierResults}
\end{table*}

\subsection{Feature Selection}
In order to evaluate the effectiveness of these five features, we used two feature selection methods: (1) information gain (a filter-based method) and (2) subset selection (a wrapper method). Table~\ref{infoGain} shows the features ranked using information gain. For example,  WordNet synonyms are the most discriminating features; whereas word length is the least.

On the other hand, using the subset selection method, the best subset of features was determined to be the frequency in the Web1T Google N-gram Corpus $+$ the number of synonyms in WordNet $+$ the psycholinguistic feature. It is interesting to note that both methods identify the psycholinguistic feature and the number of synonyms in WordNet as two of the most discriminating features; and word length as less useful.

\section{Results and Discussion}

We experimented with various learning models trained on the features described above. As a baseline, we used a Na{\"i}ve Bayes classifier. Table \ref{ClassifierResults} shows the weighted average precision, recall and F-measure on the training set evaluated using 10 fold cross-validation. As can be seen in Table \ref{ClassifierResults}, all learning models perform significantly better in classifying simple words rather than complex words. Based on the weighted average of the learning models, we submitted two Random Forest models: CLaC-EDLK-RF\_0.6 and CLaC-EDLK-RF\_0.5. The only difference between these submissions is the threshold for class assignment. In the first submission, we used the threshold of 0.5, and in the second we used 0.6. Our official ranking at the shared task ranked our CLaC-EDLK-RF\_0.6 as $21^{st}$ and CLaC-EDLK-RF\_0.5 as $24^{th}$ out of 45 systems using the G-score. However, based on the F-score, CLaC-EDLK-RF\_0.5 ranked $26^{th}$ and CLaC-EDLK-RF\_0.6 ranked $30^{th}$.

\section{Future Work}

For future work, we plan to investigate the use of other linguistic and psycholinguistic features. Because the MRC database contained a concreteness score for only 8,228 words, we plan to combine more psycholinguistic features from the same database. In addition, it would be interesting to investigate the influence of context on the readers' understanding of a target word. To do this, we could examine a window-based approach to consider the surrounding words which may affect the complexity of a word. 

\subsection*{Acknowledgement}
The authors would like to thank the anonymous reviewers for their feedback on the paper. This work was financially supported by NSERC.

\bibliographystyle{naaclhlt2016}

\end{document}